\documentclass[10pt,twocolumn,letterpaper]{article}

\usepackage{wacv}
\usepackage{times}
\usepackage{epsfig}
\usepackage{graphicx}
\usepackage{times}
\usepackage{epsfig}
\usepackage{graphicx}
\usepackage{amsmath}
\usepackage{amssymb}
\usepackage{multirow}
\usepackage[normalem]{ulem}
\usepackage{comment}
\usepackage{sidecap}
\usepackage{xcolor}
\usepackage{booktabs}
\usepackage{authblk}
\usepackage{tabularx} 
\usepackage{algorithm}
\usepackage[noend]{algpseudocode}

\usepackage[flushleft]{threeparttable}

\usepackage[pagebackref=true,breaklinks=true,letterpaper=true,colorlinks,bookmarks=false]{hyperref}

\DeclareMathOperator*{\argmin}{arg\,min}

\DeclareMathOperator{\Acc}{Acc}

\newcommand{\be}{\mathbf{e}}

\newcommand{\bx}{\mathbf{x}}

\newcommand{\sE}{\mathcal{E}}

\newcommand{\sX}{\mathcal{X}}
\newcommand{\sY}{\mathcal{Y}}
\newcommand{\sZ}{\mathcal{Z}}

\newcommand{\bbR}{\mathbb{R}}

\newcommand{\figref}[1]{Figure~\ref{#1}}
\newcommand{\eqnref}[1]{(\ref{#1})}

\newcommand{\tabref}[1]{Table~\ref{#1}}

\wacvfinalcopy
\ifwacvfinal\fi
\begin{document}

\title{Transductive Zero-Shot Learning for 3D Point Cloud Classification}


\author{\large Ali Cheraghian$^{1,2}$, Shafin Rahman$^{1,2}$, Dylan Campbell$^{1,3}$ and Lars Petersson$^{1,2}$\\
\vspace{-3mm}
\large $^1$Australian National  University, $^2$Data61-CSIRO, $^3$Australian Centre for Robotic Vision\\
{\tt\small firstname.lastname@anu.edu.au}}


\maketitle
\ifwacvfinal\thispagestyle{empty}\fi

\begin{abstract}
Zero-shot learning, the task of learning to recognize new classes not seen during training, has received considerable attention in the case of 2D image classification. However despite the increasing ubiquity of 3D sensors, the corresponding 3D point cloud classification problem has not been meaningfully explored and introduces new challenges. This paper extends, for the first time, transductive Zero-Shot Learning (ZSL) and Generalized Zero-Shot Learning (GZSL) approaches to the domain of 3D point cloud classification. To this end, a novel triplet loss is developed that takes advantage of unlabeled test data. While designed for the task of 3D point cloud classification, the method is also shown to be applicable to the more common use-case of 2D image classification. An extensive set of experiments is carried out, establishing state-of-the-art for ZSL and GZSL in the 3D point cloud domain, as well as demonstrating the applicability of the approach to the image domain.\footnote{Code  and  evaluation  protocols  available at: \url{https://github.com/ali-chr/Transductive_ZSL_3D_Point_Cloud}} 
\end{abstract}


\section{Introduction}
Capturing 3D point cloud data from complex scenes has been facilitated recently by inexpensive and accessible 3D depth camera technology. This in turn has increased the interest in, and need for, 3D object classification methods that can operate on such data. However, much if not most of the data collected will belong to classes for which a classification system may not have been explicitly trained. In order to recognize such previously ``unseen'' classes, it is necessary to develop Zero-Shot Learning (ZSL) methods in the domain of 3D point cloud classification. While such methods are typically trained on a set of so-called ``seen'' classes, they are capable of classifying certain ``unseen'' classes as well. Knowledge about unseen classes is introduced to the network via semantic feature vectors that can be derived from networks pre-trained on image attributes or on a very large corpus of texts~\cite{Hinton_NIPS_2009, Akata_PAMI_2016, Zhang_2017_CVPR, Xian_CVPR_2017}.

\begin{figure}[!t]\centering
\includegraphics[width=1\linewidth,trim=0cm 0cm 0cm 0cm, clip]{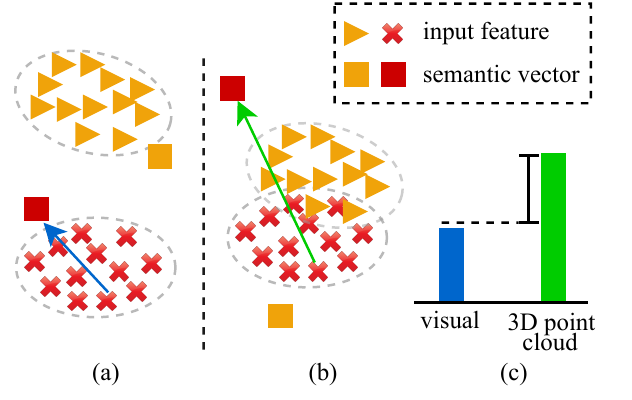}
\caption{The challenge of zero-shot learning for 3D point cloud data.
(a) and (b) are pre-trained 2D image and 3D point cloud feature spaces respectively. (c) The average intra-class distance between an unseen feature vector and a semantic feature vector of the corresponding class after inductive learning in the visual and point cloud domains respectively. The embedding space quality is much higher in (a) than (b) because image-based pre-trained models, such as ResNet, use deeper networks trained on millions of images, whereas point cloud-based models, such as PointNet, use shallower networks trained on only a few thousand point clouds.
}
\label{fig:motivation}
\vspace{-3mm}
\end{figure}
\vspace{-1mm}
Performing ZSL for the purpose of 3D object classification is a more challenging task than ZSL applied to 2D images~\cite{rahman2018unified,Akata_PAMI_2016,Changpinyo_2016_CVPR,Hinton_NIPS_2009,Lampert_PAMI_2014,Xian_CVPR_2017}. ZSL methods in the 2D domain commonly take advantage of pre-trained models, like ResNet \cite{He2016DeepRL}, that have been trained on millions of labeled images featuring thousands of classes. As a result, the extracted 2D features are very well clustered. By contrast, there is no parallel in the 3D point cloud domain; labeled 3D datasets tend to be small and have only limited sets of classes. For example, pre-trained models like PointNet \cite{Article1} are trained on only a few thousand samples from a small number of classes. This leads to poor-quality 3D features with clusters that are not nearly as well separated as their visual counterparts.
This gives rise to the problem of projection domain shift~\cite{Fu_PAMI_2015}. In essence, this means that the function learned from seen samples is biased, and cannot generalize well to unseen classes. In the \emph{inductive learning} approach, where only seen classes are used during training, projected semantic vectors tend to move toward the seen feature vectors, making the intra-class distance between corresponding unseen semantic and feature vectors large.
This intuition is visualized in Figure \ref{fig:motivation}.


Now, the key question is how far these problems can be mitigated by adopting a \emph{transductive learning} approach, where the model is trained using both labeled and unlabeled samples. Our goal is to design a strategy that reduces the bias and encourages the projected semantic vectors to align with their true feature vector counterparts, minimizing the average intra-class distance.
In 2D ZSL, the transductive setting has been shown to be effective~\cite{Fu_PAMI_2015,Zhao_NIPS_2018,Song2018TransductiveUE}, however in the case of 3D point cloud data it is a more challenging task. Pre-trained 3D features are poorly clustered and exhibit large intra-class distances. As a result, state-of-the-art transductive methods suitable for image data \cite{Fu_PAMI_2015,Zhao_NIPS_2018,Song2018TransductiveUE} are unable to reduce the bias problem for 3D data.


In order to take advantage of the transductive learning approach for 3D point cloud zero-shot learning, we propose a transductive ZSL method using a novel triplet loss that is employed in an unsupervised manner.  Unlike the traditional triplet formulation \cite{Facenet,BMVC17Zeroshot}, our proposed triplet loss works on unlabeled (test) data and can operate without the need of ground-truth supervision. This loss applies to unlabeled data such that intra-class distances are minimized while also maximizing inter-class distances, reducing the bias problem. As a result, a prediction function with greater generalization ability and effectiveness on unseen classes is learned. Moreover, our proposed method is also applicable in the case of 2D ZSL, which demonstrates the generalization strength of our method to other sensor modalities.

%

Our main contributions are:
\textbf{(1)} extending and adapting transductive zero-shot learning and generalized zero-shot learning to 3D point cloud classification for the first time;
\textbf{(2)} developing a novel triplet loss that takes advantage of unlabeled test data, applicable to both 3D point cloud data and 2D images; and
\textbf{(3)} performing extensive experiments, establishing state-of-the-art on four 3D datasets, ModelNet10~\cite{Article10}, ModelNet40~\cite{Article10}, McGill~\cite{Article49}, and SHREC2015~\cite{Article48}.

\section{Related Work}

\noindent\textbf{Zero-Shot Learning:} For the ZSL task, there has been significant progress, including on image recognition~\cite{rahman2018unified,Zhang_2017_CVPR,Akata_PAMI_2016,Changpinyo_2016_CVPR,Hinton_NIPS_2009,Lampert_PAMI_2014,Xian_CVPR_2017}, multi-label ZSL~\cite{Lee_2018_CVPR,rahman2018deep}, and zero-shot detection~\cite{rahman2018ZSD}. Despite this progress, these methods solve the constrained problem where the test instances are restricted to only unseen classes, rather than being from either seen or unseen classes. This setting, where both seen and unseen classes are considered at test time, is called Generalized Zero-Shot Learning (GZSL). To address this problem, some methods decrease the scores that seen classes produce by a constant value \cite{Chao_ECCV_2016}, while others perform a separate training stage intended to balance the probabilities of the seen and unseen classes \cite{rahman2018unified}. 
Schonfeld \etal \cite{Schonfeld_2019_CVPR} learned a shared latent space of image features and semantic representation based on a modality-specific VAE model. In our work, we use a novel unsupervised triplet loss to address the bias problem, leading to significantly better GZSL results.  

\noindent\textbf{Transductive Zero-shot Learning:} The transductive learning approach takes advantage of unlabeled test samples, in addition to the labeled seen samples. For example, 
Rohrbach \etal \cite{Rohrbach_NIPS_2013} exploited the manifold structure of unseen classes using a graph-based learning algorithm to leverage the neighborhood structure within unseen classes. 
Yu \etal \cite{Yu_TCy_2018} proposed a transductive approach to predict class labels via an iterative refining process. 
More recently, transductive ZSL methods have started exploring how to improve the accuracy of both the seen and unseen classes in generalized ZSL tasks \cite{Zhao_NIPS_2018,Song2018TransductiveUE}. Zhao \etal \cite{Zhao_NIPS_2018} proposed a domain invariant projection method that projects visual features to semantic space and reconstructs the same feature from the semantic representation in order to narrow the domain gap. In another approach, Song \etal \cite{Song2018TransductiveUE} identified the model bias problem of inductive learning, that is, a trained model assigns higher prediction scores for seen classes than unseen. To address this, they proposed a quasi-fully supervised learning method to solve the GZSL task. Xian \etal \cite{Xian_2019_CVPR} proposed f-VAEGAN-D2 which takes advantage of both VAEs and GANs to learn the feature distribution of unlabeled data. All of these approaches are designed for transductive ZSL tasks on 2D image data. In contrast, we explore to what extent a transductive ZSL setting helps to improve 3D point cloud recognition.

\noindent{\textbf{Learning with a Triplet Loss:}} Triplet losses have been widely used in computer vision \cite{Facenet,BMVC17Zeroshot,Dong_2018_ECCV,He_2018_CVPR,Do_2019_CVPR}.
Schroff \etal \cite{Facenet} demonstrated how to select positive and negative anchor points from visual features within a batch.
Qiao \etal \cite{BMVC17Zeroshot} introduced using a triplet loss to train an inductive ZSL model.
More recently, Do \etal \cite{Do_2019_CVPR} proposed a tight upper bound of the triplet loss by linearizing it using class centroids,
Zakharov \etal \cite{8202207} explored the triplet loss in manifold learning,
Srivastava \etal \cite{article1323} investigated weighting hard negative samples more than easy negatives, and
Zhaoqun \etal \cite{Li2018AngularTL} proposed the angular triplet-center loss, a variant that reduces the similarity distance between features.
%
Triplet loss related methods typically work under inductive settings, where the ground-truth label of an anchor point remains available during training. In contrast, we describe a triplet formation technique in the transductive setting. Our method utilizes test data without knowing its true label. Moreover, we choose positive and negative samples of an anchor from word vectors instead of features.

\noindent\textbf{ZSL on 3D Point Clouds:}
Despite much progress on 3D point cloud classification using deep learning~\cite{Article1,Article2,Article24,Article28,Article29,Xie_2018_CVPR,8658405,ramasinghe2019representation,ramasinghe2019spectralgans,ramasinghe2019blended}, only two works have addressed the ZSL problem for 3D point clouds.
Cheraghian \etal \cite{cheraghian2019zeroshot, cheraghian2019mitigating} proposed a bilinear compatibility function to associate a PointNet \cite{Article1} feature vector with a semantic feature vector, and separately proposed an unsupervised skewness loss to mitigate the hubness problem. Both works use inductive inference and are therefore less able to handle the bias towards seen classes in the GZSL task than our proposed method.

\section{Transductive ZSL for 3D Point Clouds}

\begin{figure}
\centering
\begin{minipage}[b]{0.38\columnwidth}
  \centering
  \centerline{\includegraphics[width=1.3\linewidth,trim=0cm 0cm 0cm 0cm, clip]{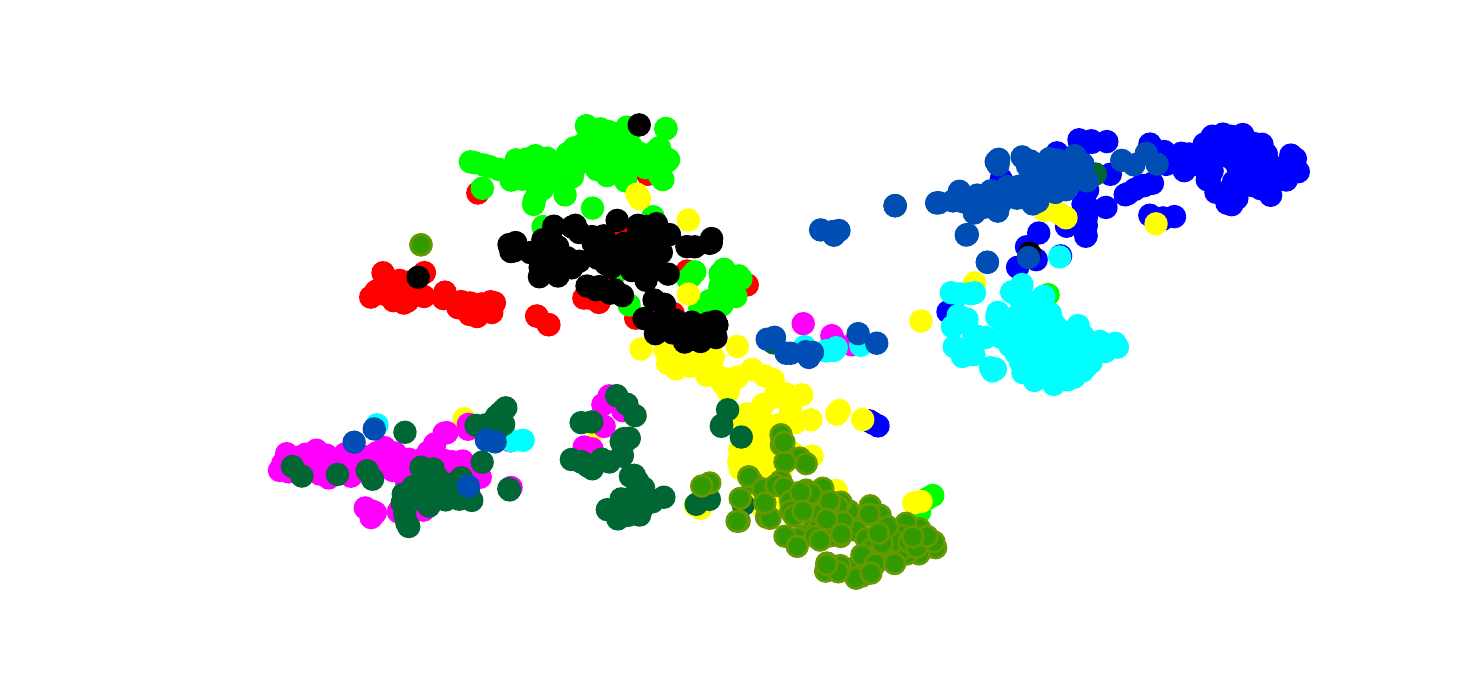}}
  \centerline{\small (a) ModelNet10}\medskip
\end{minipage}
\begin{minipage}[b]{0.5\columnwidth}
  \centering
  \centerline{\includegraphics[width=.5\linewidth,trim=0cm 0cm 0cm 0cm, clip]{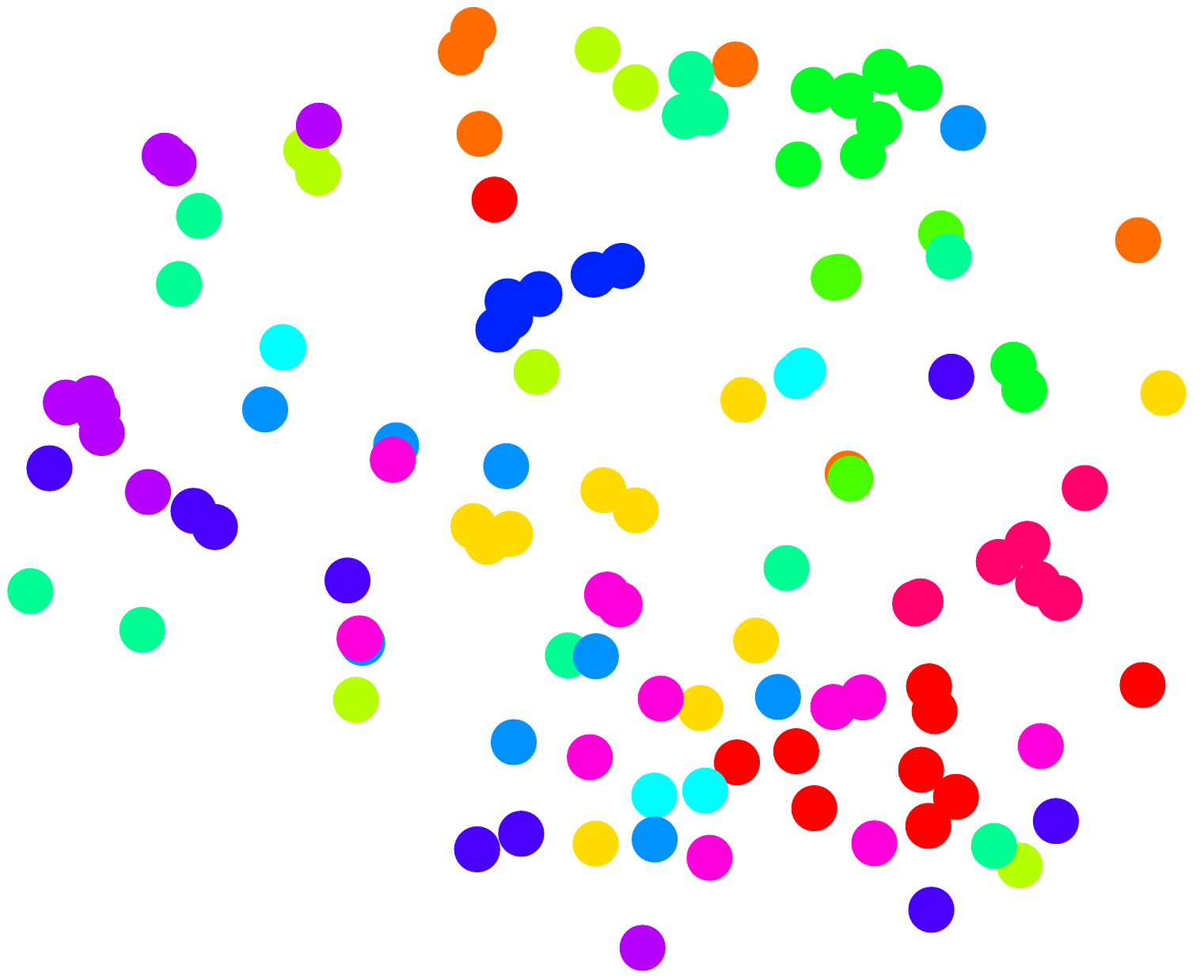}}
  \centerline{\small (b) McGill}\medskip
  \vspace{-0mm}
\end{minipage}

\vspace{-2.5mm}
\begin{minipage}[b]{0.38\columnwidth}
  \centering
  \centerline{\includegraphics[width=0.7\linewidth,trim=0cm 0cm 0cm 0cm, clip]{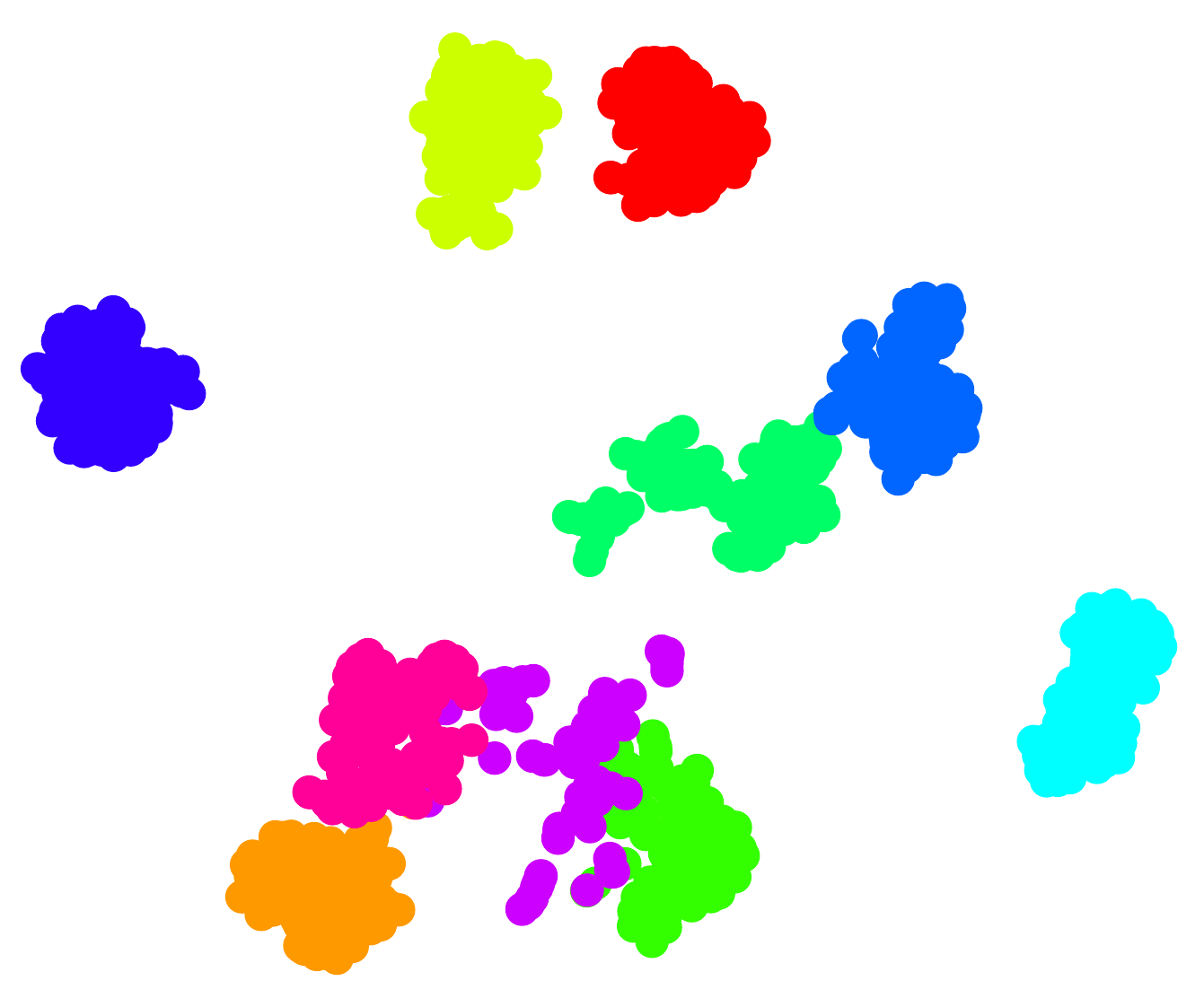}}
  \centerline{\small (c) AwA2 }\medskip
\end{minipage}
\begin{minipage}[b]{0.45\columnwidth}
  \centering
  \centerline{\includegraphics[width=.75\linewidth,trim=0cm 0cm 0cm 1.4cm, clip]{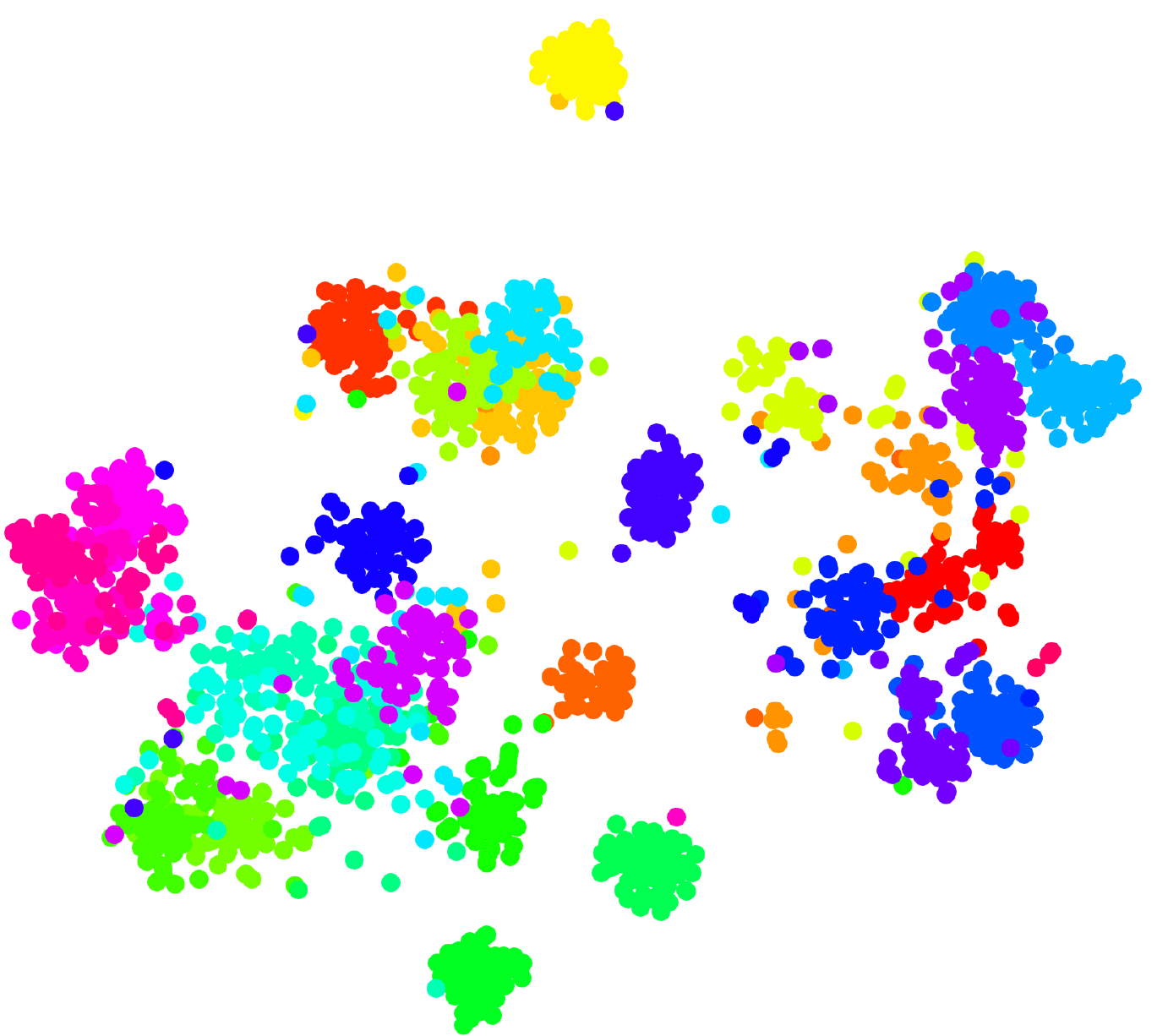}}
  \centerline{\small (d) CUB}\medskip
  
\end{minipage}


\caption{tSNE~\cite{tSNE_van2014} visualizations of unseen 3D point cloud features of (a) ModelNet10~\cite{Article10} (b) McGill~\cite{Article49} and unseen 2D image features of (c) AwA2~\cite{Xian_CVPR_2017} (d) CUB~\cite{CUB_2011} . The cluster structure in the 2D feature space is much better defined, with tighter and more separated clusters than those in the 3D point cloud.}
\label{fig:2D_versus_3D}
\end{figure}

Zero-shot learning is heavily dependent on good pre-trained models generating well-clustered features \cite{norouzi_arXiv_2013,Changpinyo_2016_CVPR,Akata_PAMI_2016,Yu_TCy_2018} as the performance of established ZSL methods otherwise degrades rapidly. In the 2D case, pre-trained models are trained by considering thousands of classes and millions of images~\cite{Xian_CVPR_2017}. However, similar quality pre-trained models are typically unavailable for 3D point cloud objects. Therefore, 3D point cloud features cluster more poorly than image features. To illustrate this point, in Figure \ref{fig:2D_versus_3D} we visualize 3D features of \emph{unseen} classes from the 3D datasets ModelNet10~\cite{Article10}, McGill~\cite{Article49} and 2D features of unseen classes from the 2D datasets AwA2~\cite{Xian_CVPR_2017} and CUB~\cite{CUB_2011}. Here, we use unseen classes to highlight the generalization ability of the pre-trained model. Because of the use of a large dataset (like ImageNet) for the 2D case, the cluster structure is more separable in 2D than in 3D. As 3D features are not as robust and separable as 2D features, relating those features to their corresponding semantic vectors is more difficult than for the corresponding 2D case. Addressing the poor feature quality of typical 3D datasets, we propose to use a triplet loss in the transductive setting of ZSL. Our method specifically addresses the alignment of poor features (like those coming from 3D feature extractors) with semantic vectors. Therefore, while our method improves the results for both 2D and 3D modalities, the largest gain is observed in the 3D case.

\subsection{Problem Formulation}
Let $\sX = \{\bx_{i}\}_{i = 1}^{n}$ for $\bx_{i}\in\bbR^{3}$ denote a 3D point cloud. Also let $\sY^{s} = \{y_{i}^{s}\}_{i = 1}^{S}$ and $\sY^{u} = \{y_{i}^{u}\}_{i = 1}^{U}$ denote disjoint ($\sY^{s}\cap\sY^{u}=0$) seen and unseen class label sets with sizes $S$ and $U$ respectively,
and $\sE^{s}=\{\phi(y^{s})\}_{i = 1}^{S}$ and $\sE^{u}=\{\phi(y^{u})\}_{i = 1}^{U}$ denote the sets of associated semantic embedding vectors for the embedding function $\phi(\cdot)$, with $\phi(y)\in\bbR^{d}$.
Then we define the set of $n_{s}$ seen instances as $\sZ^{s} = \{(\sX_{i}^{s}, l_{i}^{s}, \be_{i}^{s})\}_{i=1}^{n_{s}}$, where $\sX_{i}^{s}$ is the $i$\textsuperscript{th} point cloud of the seen set with label $l_{i}^{s} \in \sY^{s}$ and semantic vector $\be_{i}^{s} = \phi(l_{i}^{s}) \in \sE^{s}$.
The set of $n_{u}$ unseen instances is defined similarly as $\sZ^{u} = \{(\sX_{i}^{u},l_{i}^{u},\be_{i}^{u})\}_{i=1}^{n_{u}}$, where $\sX_{i}^{u}$ is the $i$\textsuperscript{th} point cloud of the unseen set with label $l_{i}^{u} \in \sY^{u}$ and semantic vector $\be_{i}^{u} = \phi(l_{i}^{u}) \in \sE^{u}$.

We consider two learning problems in this work: zero-shot learning and its generalized variant. The goal of each problem is defined as follows.
\begin{itemize}
  \item \textbf{Zero-Shot Learning (ZSL):} To predict a class label $\hat{y}^{u} \in \sY^{u}$ from the unseen label set given an unseen point cloud $\sX^{u}$.
  \item\textbf{Generalized Zero-Shot Learning (GZSL):} To predict a class label $\hat{y} \in \sY^{s}\cup\sY^{u}$ from the seen or unseen label sets given a point cloud $\sX$.
\end{itemize}

\subsection{Model Training}
Zero-shot learning can be addressed using inductive or transductive inference.
For inductive ZSL, the model is trained in a fully-supervised manner with seen instances only from the set $\sZ^{s}$.

To learn an inductive model, an objective function
\begin{align}
L_{I}=\frac{1}{N}\sum_{i=1}^{N} \left\| \varphi (\sX_{i}^{s})-\Theta (\be_{i}^{s}; W) \right \|_{2}^{2} + \lambda \left \| W \right \|_{2}^{2}
\label{eqn:LI}
\end{align}
is minimized, where $N$ is the number of instances in the batch, $\varphi(\sX_{i}^{s}) \in \bbR^{m}$ is the point cloud feature vector associated with point cloud $\sX_{i}^{s}$, $W$ are the weights of the nonlinear projection function $\Theta(\cdot)$ that maps from the semantic embedding space $\sE$ to the point cloud feature space,
and the parameter $\lambda$ controls the amount of regularization.



In contrast, transductive ZSL additionally uses the set of unlabeled, unseen instances $\{\sX_{i}^{u}\}$ and the set of unseen semantic embedding vectors $\sE^{u}$ during training.
To learn a transductive model in a semi-supervised manner, an objective function
\begin{align}
L_{T}=\frac{1}{N}\sum_{i=1}^{N}\left \| \varphi (\sX_{i}^{s})-\Theta (\be_{i}^{s}; W) \right \|_{2}^{2}
+ \alpha L_{u} + \lambda \left \| W \right \|_{2}^{2} \label{eqn:LT}
\end{align}
is minimized, where $N$ is the batch size of seen instances, $L_{u}$ is the unsupervised loss, $\alpha$ controls the influence of the unsupervised loss, and $\lambda$ controls the amount of regularization. For the $L_{u}$ term, a triplet loss is proposed, which will be outlined in the next section.

\begin{figure}[!t]\centering
\includegraphics[width=1\linewidth,trim=2cm .1cm 1cm .8cm, clip]{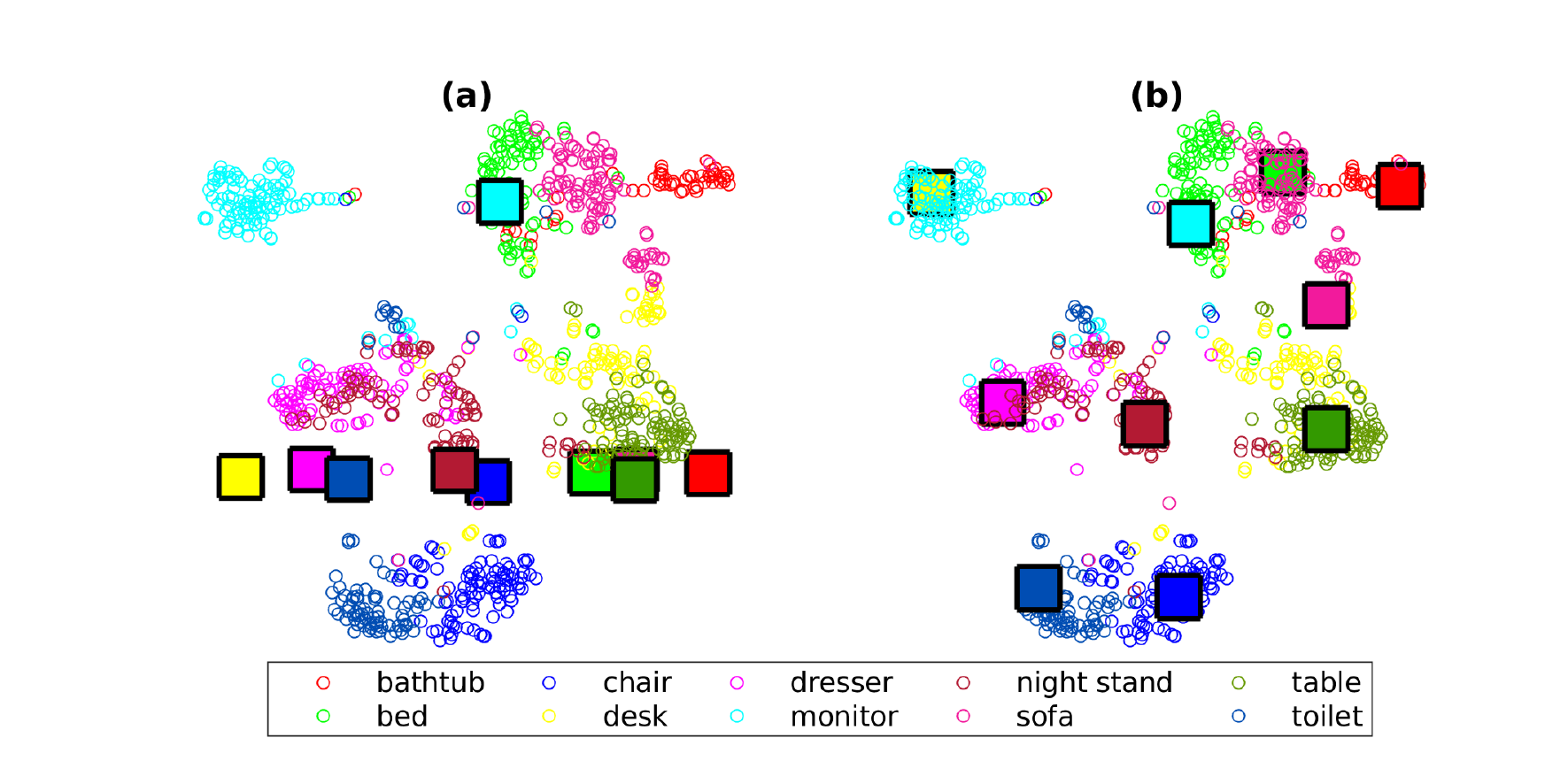}
\caption{2D tSNE~\cite{tSNE_van2014} visualization of unseen point cloud feature vectors (circles) and projected semantic feature vectors (squares) based on (a) inductive and (b) transductive learning on ModelNet10. The projected semantic feature vectors are much closer to the cluster centres of the point cloud feature vectors for transductive ZSL than for inductive ZSL, showing that the transductive approach is able to narrow the domain gap between seen and unseen classes.
}
\label{fig:inductive_transductive}
\vspace{-4mm}
\end{figure}

Transductive ZSL addresses the problem of the projection domain shift~\cite{Fu_PAMI_2015} inherent in inductive ZSL approaches. 
In ZSL, the seen and unseen classes are disjoint and often only very weakly related. Since the underlying distributions of the seen and unseen classes may be quite different, the ideal projection function between the semantic embedding space and point cloud feature space is also likely to be different for seen and unseen classes. 
As a result, using the projection function learned from only the seen classes without considering the unseen classes will cause an unknown bias.
Transductive ZSL reduces the domain gap and the resulting bias by using unlabeled unseen class instances during training, improving the generalization performance.
The effect of the domain shift in ZSL is shown in Figure~\ref{fig:inductive_transductive}.
When inductive learning is used (a), the projected unseen semantic embedding vectors are far from the cluster centres of the associated point cloud feature vectors, however when transductive learning is used (b), the vectors are much closer to the cluster centres.

\subsection{Unsupervised Triplet Loss}
In this work, we propose an unsupervised triplet loss for $L_{u}$ \eqnref{eqn:LT}. It is unsupervised because the computation of $L_{u}$ operates on test data, which remains unlabeled, and receives no ground-truth supervision throughout transductive training. To compute a triplet loss, a positive and negative sample need to be found for each anchor sample \cite{Facenet}. In the fully-supervised setting, selecting positive and negative samples is not difficult, because all training samples have ground-truth labels. However, it is much more challenging in the unsupervised setting, where ground-truth labels are not available. For transductive ZSL, we define a positive sample using a \emph{pseudo-labeling} approach \cite{Pseudo_label}. 
For each anchor $\sX^{u}$, we assign a pseudo-label that chooses a positive sample $\be^{+}$ 
among the semantic embedding vectors which is the closest to the anchor feature vector $\varphi(\sX^{u})$ after projection $\Theta(\cdot)$, as follows
\begin{align}
\be^{+} = \argmin_{\be \in \sE^{u}} \| \varphi (\sX^{u})-\Theta (\be; W) \|_{2} \textrm{.}\label{arg_ZSL}
\end{align}

Such pseudo-labeling is different from the usual practice \cite{Pseudo_label} because it chooses a semantic vector as a positive sample in the triplet formation instead of a plausible ground-truth label. For GZSL, the unlabeled data $\sX^{c}$ for $c \in \{s, u\}$ can be from the seen or unseen classes during training. As a result, a pseudo-label must be found for both unlabeled seen and unlabeled unseen samples.
Importantly, if the pseudo-label indicates that an unlabeled sample is from a seen class, then that sample is discarded. This reduces the impact of incorrect, noisy pseudo-labels on the model for seen classes. Samples from seen classes (with ground-truth labels) will instead influence the supervised loss function. Hence, we use true supervision where possible (seen classes), and only use pseudo-supervision where there is no alternative (unseen classes).
The positive sample for GZSL is therefore chosen as follows
\begin{align}
\be^{+} = \argmin_{\be \in \sE^{s}\cup \sE^{u}} \| \varphi (\sX^{c})-\Theta (\be; W) \|_{2}.\label{arg_GZSL}
\end{align}

The negative sample is selected from the seen semantic embedding set $\sE^{s}$ for both ZSL and GZSL, since all elements of this set will have a different label from the unseen anchor. 
We choose the negative sample as the seen semantic embedding vector whose projection is closest to the anchor vector $\varphi (\sX^{u})$,
\begin{align}
\be^{-} = \argmin_{\be \in \sE^{s}} \| \varphi (\sX^{s})-\Theta (\be; W) \|_{2}\label{arg_neg}
\end{align}


Finally, the unsupervised loss function $L_{u}$ associated with the unlabeled instances for both ZSL and GZSL tasks is defined as follows:
\begin{align}
L_{u} =\frac{1}{N^{\prime}}\sum_{i=1}^{N^{\prime}} &\max \bigg\{ 0,\left \| \varphi (\sX_{i}^{u})-\Theta(\be^{+}; W) \right \|_{2}^{2}\nonumber\\
&\hspace{5pt} + m - \left \| \varphi (\sX_{i}^{u})-\Theta(\be^{-}; W) \right \|_{2}^{2} \bigg\}
\end{align}
where $m$ is a margin that encourages separation between the clusters, and $N^{\prime}$ is the batch size of the unlabeled instances. We describe the overall training process in Algorithm \ref{alg:method}. In the proposed algorithm, in the first stage, an inductive model $W_{ind}$ is learned. Then the transductive model $W_{tns}$ is initialized with the inductive model. Finally the transductive model is learned.  

This proposed triplet loss is distinct from recent literature \cite{Facenet,BMVC17Zeroshot} in two ways. \textbf{(1)} Popular methods of triplet formation select a similar feature to the input feature as a positive sample, whereas we choose a semantic word vector for this purpose. This helps to better align the 3D point cloud features with the semantic vectors. \textbf{(2)} We employ a triplet loss in a transductive setting to utilize unlabeled (test) data, whereas established methods consider the triplet loss for inductive training only. This extends the role of the triplet loss beyond inductive learning.

%

\begin{algorithm}[!t]
\caption{Transductive ZSL for 3D point cloud objects}\label{euclid}
\begin{algorithmic}[1]
\Statex \textbf{Input:} $\sX^{s}$, $\sY^{s}$, $\sE^{s}$, $n_{s}$, $\sX^{u}$, $\sE^{u}$, $n_{u}$
\Statex \textbf{Output:} A trained model $W_{tns}$ to find $\hat{y}$ for all $\sX^{u}$

\Statex \textbf{Inductive training stage}\\
$W_{ind} \gets$ train an inductive model using Eq~\ref{eqn:LI} with only seen data: $\sX^{s}$, $\sY^{s}$, $\sE^{s}$, $n_{s}$

\Statex \textbf{Transductive training stage}\\
$W_{tns} \gets W_{ind}$, initialize transductive model

\Repeat
\If{GZSL}
\State  $\hat{y} \gets$ use $W_{tns}$ to assign positive and negative anchors to $\sX^{u}$ using Eq~\ref{arg_GZSL} and Eq~\ref{arg_neg} for triple formation
\Else
\State  $\hat{y} \gets$  use $W_{tns}$ to assign  positive and negative anchors to $\sX^{u}$ using Eq~\ref{arg_ZSL} and Eq~\ref{arg_neg} for triple formation
\EndIf

\For{$\forall I \in \sX^{s} \cup   \sX^{u}$}
         \State Calculate overall transductive loss using Eq~\ref{eqn:LT}
         \State Backpropagate and update $W_{tns}$
\EndFor

\Until convergence

\Statex \textbf{Return} Class decision $\hat{y}$ with $W_{tns}$ using Eq~\ref{eqn:inference_ZSL} for ZSL or Eq~\ref{eqn:inference_GZSL} for GZSL

\end{algorithmic}
\label{alg:method}
\end{algorithm}


\subsection{Model Architecture}

\begin{figure}
\centering
\includegraphics[width=.9\linewidth,trim=0cm 0cm 0cm 0cm, clip]{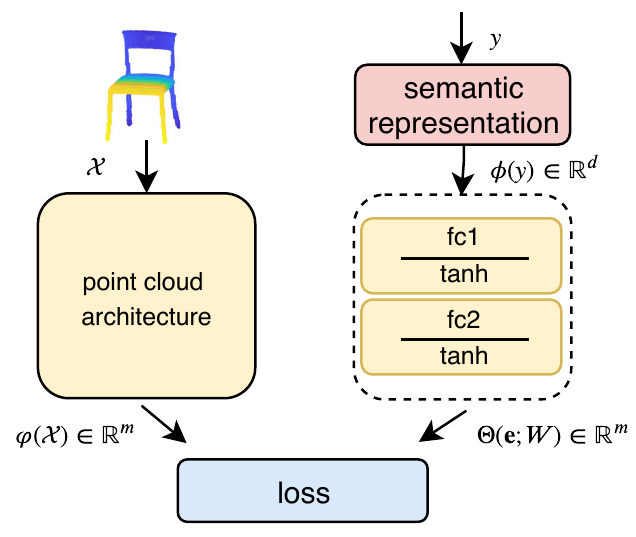}
\caption{The proposed architecture for ZSL and GZSL. For inductive learning, the input point cloud and semantic representation are $\sX = \sX^{s}$ and $\be = \phi(y) \in \sE^{s}$, respectively. For transductive learning, the input point cloud and semantic representation are $\sX = \sX^{s} \cup \sX^{u}$ and $\be \in \sE^{s} \cup \sE^{u}$ respectively.}
\label{fig:model_architecture}
\end{figure}

The proposed model architecture is shown in Figure~\ref{fig:model_architecture}, consisting of two branches: the point cloud network that extracts a feature vector $\varphi(\sX) \in \bbR^{m}$ from a point cloud $\sX$, and the semantic projection network that projects a semantic feature vector $\be\in \bbR^{d}$ into point cloud feature space.
Any network that learns a feature space from 3D point sets and is invariant to permutations of points in the point cloud can be used in our method as the point cloud network \cite{Article1,Article2,Article24,Article27,Article28,Article29,Xie_2018_CVPR}. The projection network $\Theta(\cdot)$ with trainable weights $W$ consists of two fully-connected layers, with $512$ and $1024$ dimensions respectively, each followed by a $\tanh$ nonlinearity.

\subsection{Inference}

For the zero-shot learning task, given the learned optimal weights $W$ from training with labeled seen instances $\sX^{s}$ and unlabeled unseen instances $\sX^{u}$, the label of the input point cloud $\sX^{u}$ is predicted as
\begin{align}
\hat{y} = \argmin_{y \in \sY^{u}} \left\| \varphi(\sX^{u})-\Theta(\phi(y); W) \right\|_2 \textrm{.} \label{eqn:inference_ZSL}
\end{align}
For the generalized zero-shot learning task, the label of the input point cloud $\sX^{c}$ for $c \in \{s, u\}$ is predicted as
\begin{align}
\hat{y} = \argmin_{y \in \sY^{s} \cup \sY^{u}} \left\| \varphi(\sX^{c})-\Theta(\phi(y); W) \right\|_2 \textrm{.} \label{eqn:inference_GZSL}
\end{align}

\section{Results}
\subsection{Experimental Setup}

\noindent\textbf{Datasets:} We evaluate our approach on four well-known 3D datasets, ModelNet10~\cite{Article10}, ModelNet40~\cite{Article10}, McGill~\cite{Article49}, and SHREC2015~\cite{Article48}, and two 2D datasets, AwA2~\cite{Xian_CVPR_2017} and CUB~\cite{CUB_2011}. The dataset statistics as used in this work are given in \tabref{Table:splitting}.
For the 3D datasets, we follow the seen/unseen splits proposed by Cheraghian \etal \cite{cheraghian2019zeroshot}, where the seen classes are those $30$ in ModelNet40 that do not occur in ModelNet10, and the unseen classes are those from the test sets of ModelNet10, McGill and SHREC2015 that are not in the set of seen classes. These splits allow us to test unseen classes from different distributions than that of the seen classes.
For the 2D datasets, we follow the Standard Splits (SS) and Proposed Splits (PS) of Xian \etal~\cite{Xian_CVPR_2017}.

\noindent\textbf{Semantic features:} We use the 300-dimensional word2vec~\cite{Mikolov_NIPS_2013} semantic feature vectors for the 3D dataset experiments, the 85-dimensional attribute vectors from Xian \etal~\cite{Xian_CVPR_2017} for the AwA2 experiments, and the 312-dimensional attribute vectors from Wah \etal~\cite{CUB_2011} for the CUB experiments.

\noindent\textbf{Evaluation:} We report the top-$1$ accuracy as a measure of recognition performance, where the predicted label (the class with minimum distance from the test sample) must match the ground-truth label to be considered a successful prediction. For generalized ZSL, we also report the Harmonic Mean (HM) \cite{Xian_CVPR_2017}  of the accuracy of the seen and unseen classes, computed as
\begin{align}
\textrm{HM} = \frac{2 \times \Acc_{s} \times \Acc_{u} }{\Acc_{s} + \Acc_{u}}
\end{align}
\noindent
where $\Acc_{s}$ and $\Acc_{u}$ are seen and unseen class top-$1$ accuracies respectively. 

\noindent\textbf{Cross-validation:}
We used Monte Carlo cross-validation to find the best hyper-parameters, averaging over $10$ repetitions. For ModelNet40, $17\%$ (5) of the 30 seen classes were randomly selected as an unseen validation set, while $20\%$ were used for the AwA2 and CUB datasets. The hyper-parameters $\alpha$ and $\lambda$ were 0.15 and 0.0001 for ModelNet40, 0.1 and 0.001 for AwA2, and 0.25 and 0.001 for CUB.

\noindent\textbf{Implementation details:}
For the 3D data experiments, we used PointNet \cite{Article1} as the point cloud feature extraction network, with five multi-layer perceptron layers (64,64,64,128,1024) followed by max-pooling layers and two fully-connected layers (512,1024). Batch normalization (BN)~\cite{Article45} and ReLU activations were used for each layer. The 1024-dimensional input feature embedding was extracted from the last fully-connected layer. The network was pre-trained on the 30 seen classes of ModelNet40.
For the 2D data experiments, we used a 101-layered ResNet architecture \cite{He2016DeepRL}, where the 2048-dimensional input feature embedding was obtained from the top-layer pooling unit. The network was pre-trained on ImageNet 1K \cite{imagenet_cvpr09} without fine-tuning.
We fixed the pre-trained weights for both the 3D and 2D networks.
For semantic projection layers, we used two fully-connected  (512,1024) with tanh non-linearities. These parameters are fully-learnable.
To train the network, we used the Adam optimizer~\cite{Article40} with an initial learning rate of 0.0001, and batch sizes of 32 and 128 for 3D and 2D experiments respectively.
We implemented the architecture using TensorFlow~\cite{Article46} and trained and tested it on a NVIDIA GTX Titan~V GPU.

\begin{table}[!t]\centering\small
\newcolumntype{C}{>{\centering\arraybackslash}X}
\renewcommand*{\arraystretch}{1.1}
\setlength{\tabcolsep}{2pt}
\begin{tabularx}{\columnwidth}{l l C C c}\hline
&\multirow{2}{*}{Dataset}    & Total   & Seen/   & Train/ \\
                            & &classes  & Unseen  & Valid/Test \\ \hline
\multirow{4}{*}{3D}&ModelNet40 \cite{Article10} & 40 & 30/--         & 5852/1560/--\\
&ModelNet10 \cite{Article10} & 10 & --/10         & --/--/908\\
&McGill \cite{Article49}     & 19 & --/14         & --/--/115\\
&SHREC2015 \cite{Article48}  & 50 & --/30         & --/--/192\\\hline
\multirow{4}{*}{2D}&AwA2 SS \cite{Xian_CVPR_2017} & 50 & 40/10      &  30337/--/6985\\
&AwA2 PS \cite{Xian_CVPR_2017} & 50 & 40/10      &  23527/5882/7913\\
&CUB SS \cite{CUB_2011} & 200 & 150/50           & 8855/--/2933\\
&CUB PS \cite{CUB_2011} & 200 & 150/50           & 7057/1764/2967\\\hline
\end{tabularx}
\vspace{.2em}
\caption{Statistics of the 3D and 2D datasets. The total number of classes
in the datasets are reported, alongside the actual splits used in this paper dividing the classes into seen or unseen and the elements into those used for training or testing. The 3D splits are from \cite{cheraghian2019zeroshot} and the 2D Standard Splits (SS) and Proposed Splits (PS) are from Xian \etal \cite{Xian_CVPR_2017}.}
\label{Table:splitting}
\end{table}
\subsection{3D Point Cloud Experiments}

For the experiments on 3D data, we compare with two 3D ZSL methods, ZSLPC \cite{cheraghian2019zeroshot} and MHPC \cite{cheraghian2019mitigating}, and three 2D ZSL methods, f-CLSWGAN \cite{Xian_2018_CVPR}, CADA-VAE \cite{Schonfeld_2019_CVPR}, and QFSL \cite{Song2018TransductiveUE}. These state-of-the-art image-based methods were re-implemented and adapted to point cloud data to facilitate comparison.
We also report results for a baseline inductive method, which uses the inductive loss function $L_I$ \eqnref{eqn:LI} and is trained only on labeled seen classes, and for a transductive baseline method, which replaces our triplet unlabeled loss $L_{u}$ with a standard Euclidean loss.

The results on the ModelNet10, McGill, and SHREC2015 datasets are shown in \tabref{table:ZSL_3D}. Our method significantly outperforms the other approaches on these datasets. Several observations can be made from the results.
\textbf{(1)} Transductive learning is much more effective than inductive learning for point cloud ZSL. This is likely due to inductive approaches being more biased towards seen classes, while transductive approaches alleviate the bias problem by using unlabeled, unseen instances during training.
\textbf{(2)} Although generative methods \cite{Xian_2018_CVPR, Schonfeld_2019_CVPR} have shown successful results on 2D ZSL, they fail to generalize to 3D ZSL. We hypothesize that they rely more strongly on high quality pre-trained models and attribute embeddings, both of which are not available for 3D data.
\textbf{(3)} Our proposed method performs better than QFSL, which is likely due to our triplet loss formulation. While noisy, the positive and negative samples of unlabeled data provide useful supervision, unlike the unsupervised approach for only unlabeled data in QFSL.
\textbf{(4)} The triplet loss performs much better than the Euclidean loss for this problem, since it maximizes the inter-class distance as well as minimizing the intra-class distance.
\textbf{(5)} Our proposed method does not perform as well on the McGill and SHREC2015 datasets when compared to the ModelNet10 results, because the distributions of semantic feature vectors in the unseen McGill and SHREC2015 datasets are significantly different from the distribution in the seen ModelNet40 dataset, much more so than that of ModelNet10 \cite{cheraghian2019zeroshot}.

\begin{table}[!t]\centering
\newcolumntype{C}{>{\centering\arraybackslash}X}
\renewcommand*{\arraystretch}{1.2}
\setlength{\tabcolsep}{1pt}
\begin{tabularx}{\columnwidth}{c l C C c}\hline
 &  Method &  ModelNet10 &  McGill & SHREC2015 \\\hline
\multirow{5}{*}{I} 
 & ZSLPC \cite{cheraghian2019zeroshot} & 28.0 & 10.7 & 5.2 \\ 
& MHPC \cite{cheraghian2019mitigating} & 33.9 & 12.5 & 6.2 \\ 
 & f-CLSWGAN \cite{Xian_2018_CVPR} & 20.7 & 10.2 & 5.2 \\ 
 & CADA-VAE \cite{Schonfeld_2019_CVPR} & 23.0 & 10.7 & 6.2 \\ 
& Baseline & 23.5 & 13.0 & 5.2 \\
 
 \hline
\multirow{3}{*}{T} & QFSL \cite{Song2018TransductiveUE}  & 38.8  & 18.8 & 9.5 \\
 & Baseline & 37.8 & \textbf{21.7} & 5.2 \\
 & Ours & \textbf{46.9} & \textbf{21.7} & \textbf{13.0} \\\hline
\end{tabularx}

\caption{ZSL results on the 3D ModelNet10 \cite{Article10}, McGill \cite{Article49}, and SHREC2015 \cite{Article48} datasets. We report the top-1 accuracy (\%) for each method. 
``I'' and ``T'' denote inductive and transductive learning respectively.}
\label{table:ZSL_3D}
\vspace{2mm}
\end{table}


\begin{figure}
\centering
\includegraphics[width=\columnwidth,trim=27pt 0pt 25pt 12pt, clip]{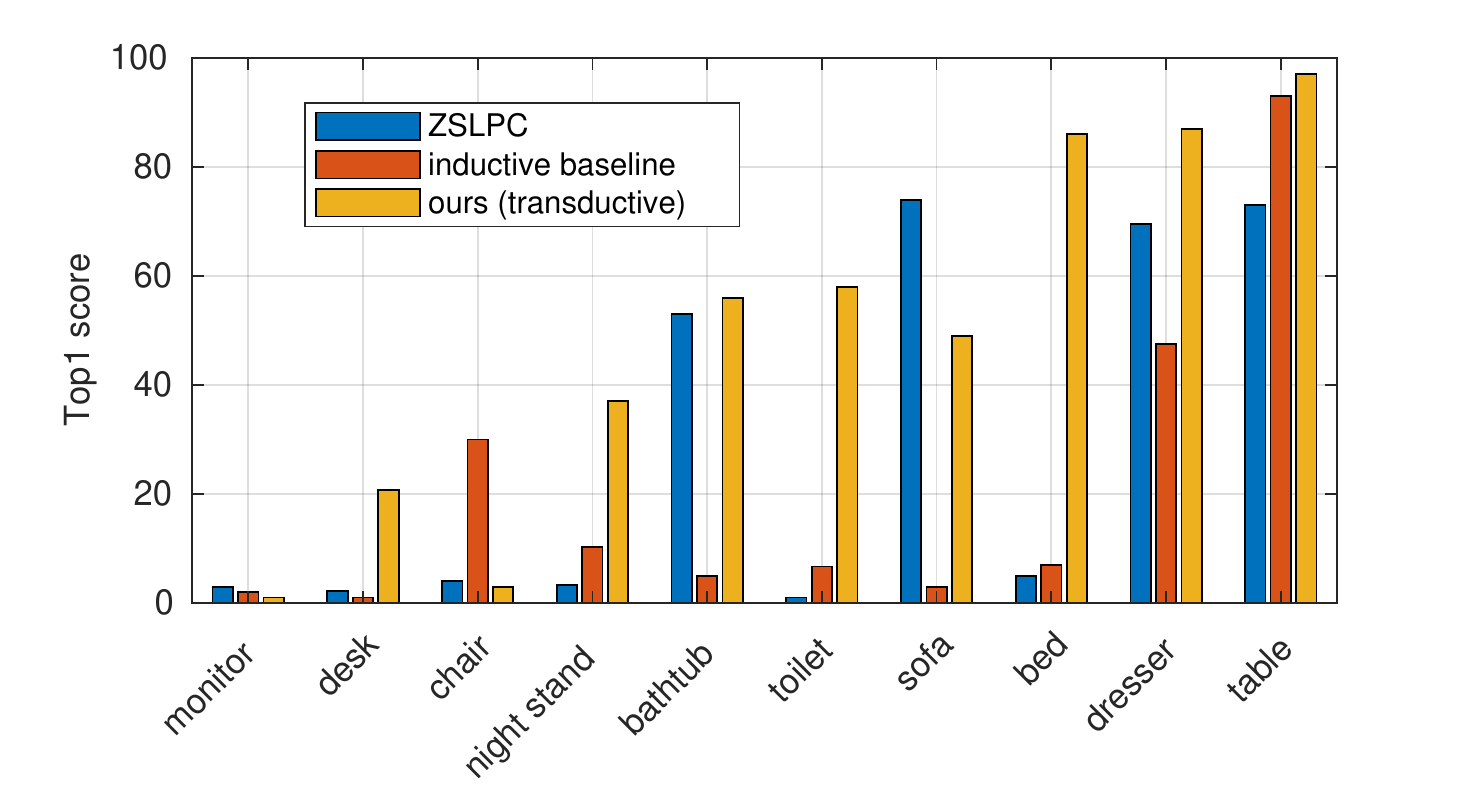}
\caption{Individual performance on unseen classes from ModelNet10. Our transductive method consistently outperforms both ZSLPC \cite{cheraghian2019zeroshot} and the  inductive baseline.}
\label{fig:indvidual_class}
\end{figure}


Generalized ZSL, which is more realistic than standard ZSL, is more challenging than ZSL as there are both seen and unseen classes during inference. As a result, methods proposed for ZSL do not usually report results for GZSL. The results are shown in \tabref{table:GZSL_3D}. Our method obtained the best performance with respect to the harmonic mean (HM) on all datasets, and the best performance with respect to the unseen class accuracy $\Acc_{u}$ on most datasets, which demonstrates the utility of our method for GZSL as well as ZSL for 3D point cloud recognition.

\begin{table*}[!t]\centering
\newcolumntype{C}{>{\centering\arraybackslash}X}
\setlength{\tabcolsep}{4pt}
\begin{tabularx}{\textwidth}{c l C C C C C C C C C}\hline
\multirow{2}{*}{} & \multirow{2}{*}{Method} & \multicolumn{3}{c}{ModelNet10} & \multicolumn{3}{c}{McGill} & \multicolumn{3}{c}{SHREC2015} \\
 &  & $\Acc_{s}$ & $\Acc_{u}$ & HM & $\Acc_{s}$ & $\Acc_{u}$ & HM & $\Acc_{s}$ & $\Acc_{u}$ & HM \\ \hline
\multirow{4}{*}{I} 
 & MHPC \cite{cheraghian2019mitigating} & 53.8& \textbf{26.2}& 35.2 & - &- & - & - & - & - \\
  & f-CLSWGAN  \cite{Xian_2018_CVPR} & 76.3& 3.7 & 7.0 & 75.3 & 2.3 & 4.5 & 74.2 & 0.8 & 1.6 \\
 & CADA-VAE \cite{Schonfeld_2019_CVPR} & \textbf{84.7} & 1.3 & 2.6 & \textbf{83.3} & 1.6 & 3.1 & 80.0 & 1.7 & 3.3 \\
 & Baseline & 83.7 & 0.4 & 0.8 & {80.0} & 0.9 & 1.8 & 82.1 & 0.9 & 1.8 \\ 
\hline
\multirow{3}{*}{T} & QFSL \cite{Song2018TransductiveUE} & 58.1 &  21.8 & 31.7 & 65.3 & 13.0 & 21.6 & 72.3 & 7.8 & 14.1 \\
 & Baseline & 77.7 & 21.0 & 33.1 & 75.5 & 12.2 & 21.0 & \textbf{83.4} & 4.2 & 8.0 \\
 & Ours & 74.6 & 23.4 & \textbf{35.6} & 74.4 & \textbf{13.9} & \textbf{23.4} & 78.6 & \textbf{10.6} & \textbf{18.4} \\ \hline
\end{tabularx}
\vspace{-.2em}
\caption{GZSL results on the 3D ModelNet10 \cite{Article10}, McGill \cite{Article49}, and SHREC2015 \cite{Article48} datasets. We report the top-1 accuracy (\%) on seen classes ($\Acc_s$) and unseen classes ($\Acc_u$) for each method, as well as the harmonic mean (HM) of both measures. ``I'' and ``T'' denote inductive and transductive learning respectively.}
\label{table:GZSL_3D}
\vspace{-1.5em}
\end{table*}

We also show, in \figref{fig:indvidual_class}, the performance of individual classes from ModelNet10. Our method achieves the best accuracy on most classes, while the inductive baseline and ZSLPC \cite{cheraghian2019zeroshot} have close to zero accuracy on many classes (\eg, desk, night stand, toilet, and bed). This is likely due to the hubness problem, which inductive methods are more sensitive to than transductive methods.

\subsection{2D Image Experiments}

While our method was designed to address ZSL and GZSL tasks for 3D point cloud recognition, we also adapt and evaluate our method for the case of 2D image recognition. The results for ZSL and GZSL are shown in Tables~\ref{table:ZSL_2D} and \ref{table:GZSL_2D} respectively. 

For ZSL, our proposed method is evaluated on the AwA2~\cite{Xian_CVPR_2017} and CUB~\cite{CUB_2011} datasets using the SS and PS splits \cite{Xian_CVPR_2017}.
Our method achieves very competitive results on these datasets, indicating that the method can generalize to image data.
Note that we do not fine-tune the image feature extraction network in our model, unlike the models listed with asterisks, for fair comparison with existing work. However, the literature demonstrates that fine-tuning can improve performance considerably, particularly on the CUB dataset.

For GZSL, we evaluate our method on the same datasets and compare with state-of-the-art GZSL methods \cite{Socher_NIPS_2013,Chao_ECCV_2016,Zhao_NIPS_2018,Song2018TransductiveUE}. As shown in \tabref{table:GZSL_2D}, our method is again competitive with the other methods on the AwA2 dataset with respect to both unseen class accuracy and harmonic mean accuracy. Our results lag state-of-the-art on the CUB dataset, although fine-tuning the feature extraction network may go some way to closing this gap.


\begin{table}[!t]\centering
\newcolumntype{C}{>{\centering\arraybackslash}X}
\renewcommand*{\arraystretch}{1.2}
\setlength{\tabcolsep}{4pt}
\begin{tabularx}{\columnwidth}{c l C C C C}\hline
\multirow{2}{*}{} & \multirow{2}{*}{Method} & \multicolumn{2}{c}{AwA2} & \multicolumn{2}{c}{CUB} \\
 &  & SS & PS & SS & PS \\ \hline
\multirow{7}{*}{I} 
 & SJE \cite{Akata_CVPR_2015} & 69.5 & 61.9 & 55.3 & 53.9 \\
 & ESZSL \cite{romera_ICML_2015} & 75.6 & 58.6 & 55.1 & 53.9 \\
 & SYNC \cite{Changpinyo_2016_CVPR} & 71.2 & 46.6 & 54.1 & 55.6 \\ 
 & f-CLSWGAN  \cite{Xian_2018_CVPR} & - & - & - & 57.3 \\ 
 & f-VAEGAN-D2 \cite{Xian_2019_CVPR} & - & 71.1 & - & 61.0 \\
& f-VAEGAN-D2{*} \cite{Xian_2019_CVPR} & - & 70.3 & - & 72.9 \\
 & Baseline & 71.2 & 69.0 & 59.3 & 54.2 \\ 
 \hline
\multirow{6}{*}{T} & DIPL \cite{Zhao_NIPS_2018} & - & - & 68.2 & 65.4 \\
& QFSL{*} \cite{Song2018TransductiveUE} & 84.8 & 79.7 & 69.7 & {72.1} \\
 & f-VAEGAN-D2 \cite{Xian_2019_CVPR} & - & \textbf{89.8} & - & 71.1 \\
  & f-VAEGAN-D2{*} \cite{Xian_2019_CVPR} & - & 89.3 & - & \textbf{82.6} \\
& Baseline & 83.3 & 75.6 & 70.6 & 58.3 \\
 & Ours & \textbf{88.1} & {87.3} & \textbf{72.0} & 62.2 \\ \hline
\end{tabularx}
\vspace{.2em}
\caption{ZSL results on the Standard Splits (SS) and Proposed Splits (PS) of the 2D AwA2 and CUB datasets. We report the top-1 accuracy (\%) for each method. ``I'' and ``T'' denote inductive and transductive learning respectively. $^{*}$Image feature extraction model fine-tuned (we do not fine-tune our model).}
\label{table:ZSL_2D}
\vspace{-3mm}

\end{table}

\begin{table}[!t]\centering
\newcolumntype{C}{>{\centering\arraybackslash}X}
\renewcommand*{\arraystretch}{1.2}
\setlength{\tabcolsep}{1pt}
\begin{tabularx}{\columnwidth}{c l C C C C C C}\hline
\multirow{2}{*}{} & \multirow{2}{*}{Method} & \multicolumn{3}{c}{AwA2} & \multicolumn{3}{c}{CUB} \\
 &  & $\Acc_{s}$ & $\Acc_{u}$ & HM & $\Acc_{s}$ & $\Acc_{u}$ & HM \\ \hline
\multirow{7}{*}{I} & CMT\cite{Socher_NIPS_2013} & 89.0 & 8.7 & 15.9 & 60.1 & 4.7 & 8.7 \\
 & CS\cite{Chao_ECCV_2016} & 77.6 & 45.3 & 57.2 & 49.4 & 48.1 & 48.7 \\ 
 & f-CLSWGAN  \cite{Xian_2018_CVPR} & - & - & - & 43.7  & 57.7 & 49.7\\
 & CADA-VAE \cite{Schonfeld_2019_CVPR} & 75.0  & 55.8 & 63.9 & 53.5 & 51.6 & 52.6 \\
  & f-VAEGAN-D2 \cite{Xian_2019_CVPR} & 57.6 & 70.6 & 63.5 &48.4 & 60.1 &53.6 \\
   & f-VAEGAN-D2{*} \cite{Xian_2019_CVPR} & 57.1 & 76.1 &65.2  & 63.2& 75.6 & 68.9\\
 & Baseline & 88.9 & 22.1 & 35.4 & 69.4 & 8.4 & 14.9\\ \hline
\multirow{6}{*}{T} & DIPL\cite{Zhao_NIPS_2018} & - & - & - & 44.8 & 41.7 & 43.2 \\
 & QFSL{*}\cite{Song2018TransductiveUE} & \textbf{93.1} & 66.2 & 77.4 & \textbf{74.9} & {71.5} & {73.2} \\
  & f-VAEGAN-D2 \cite{Xian_2019_CVPR} & 84.8 & 88.6 & 86.7 & 61.4 & 65.4 & 63.2\\
  & f-VAEGAN-D2{*} \cite{Xian_2019_CVPR} & 86.3 & \textbf{88.7} & \textbf{87.5} & 73.8&\textbf{81.4}  &\textbf{77.3} \\
 & Baseline & 88.0 & 67.2 & 76.2 & 51.4 & 40.2 & 45.1 \\
 & Ours & 81.8 & {83.1} & {82.4} & 50.5 & 50.2 & 50.3 \\ \hline
\end{tabularx}
\vspace{.2em}
\caption{GZSL results on the 2D AwA2 and CUB datasets. We report the top-1 accuracy (\%) on seen classes ($\Acc_s$) and unseen classes ($\Acc_u$) for each method, as well as the harmonic mean (HM) of both measures. ``I'' and ``T'' denote inductive and transductive learning respectively. $^{*}$Image feature extraction model fine-tuned (we do not fine-tune our model).}
\label{table:GZSL_2D}
\vspace{-1em}
\end{table}


\subsection{Discussion}

\noindent\textbf{Challenges with 3D data:} Recent deep learning methods for classifying point cloud objects have achieved over 90\% accuracy on several standard datasets, including ModelNet40 and ModelNet10. Moreover, due to significant progress in depth camera technology ~\cite{rs10020328,Izadi_3D_2011}, it is now possible to capture 3D point cloud objects at scale much more easily. It is therefore likely that many classes of 3D objects will not be present in the labeled training set. As a result, zero-shot classification systems will be needed to leverage other more easily-obtainable sources of information in order to classify unseen objects.
However, we observe that the difference in accuracy between ZSL and supervised learning is still very large for 3D point cloud classification, \eg 46.9\% as compared to 95.7\% \cite{Article27} for ModelNet10. As such, there is significant potential for improvement for zero-shot 3D point cloud classification. While the performance is still quite low, this is also the case for 2D ZSL, with state-of-the-art being 31.1\% top-5 accuracy on the ImageNet2010/12~\cite{ILSVRC_2015} datasets, reflecting the challenging nature of the problem.


\noindent\textbf{Hubness:} 
ZSL methods either (a) map the input feature space to semantic space using a hinge loss or least mean squares loss \cite{Frome_NIPS_2013,Socher_NIPS_2013}, (b) map both spaces to an intermediate space using a binary cross entropy or a hinge loss \cite{Article50,Article51}, or (c) map the semantic space to the input feature space \cite{Zhang_2017_CVPR}.
We use the last approach, projecting semantic vectors to input feature space, since it has been shown that this alleviates the hubness problem \cite{Shigeto_Hubness_2015,Zhang_2017_CVPR}.
%
We validate this claim by measuring the skewness of the distribution $N_k$ \cite{Shigeto_Hubness_2015,Article57} when projected in each direction, and the associated accuracy. We report these values in \tabref{table:hubness} for the ModelNet10 dataset. The degree of skewness is much lower when projecting the semantic feature space to the point cloud feature space, and achieves a significantly higher accuracy. This provides additional evidence that this projection direction is preferable for mitigating the problem of hubs and the consequent bias.

\begin{table}[!t]\centering
\newcolumntype{C}{>{\centering\arraybackslash}X}
\renewcommand*{\arraystretch}{1.2}
\begin{tabularx}{\columnwidth}{l C C}\hline
 $N_k\textrm{-skewness}$ & Semantic space & Input space $\rightarrow$ \\
  (Accuracy) &  $\rightarrow$ input space & semantic space\\ \hline
Inductive  & 2.67 (23.5\%) & 3.07 (19.5\%)\\
Transductive  & -0.19 (46.9\%) & 2.03 (31.2\%)\\
\hline
\end{tabularx}
\vspace{.5em}
\caption{The skewness (and accuracy) on ModelNet10 with different projection directions in both inductive and transductive settings. The skewness is lower when projecting the semantic space to the input point cloud feature space, mitigating the hubness problem and leading to more accurate transductive ZSL.}
\label{table:hubness}
\vspace{-1em}
\end{table}

\section{Conclusion}

In this paper, we identified and addressed issues that arise in the inductive and transductive settings of zero-shot learning and its generalized variant when applied to the domain of 3D point cloud classification. We observed that in the 2D domain the embedding quality generated by the pre-trained feature space is of a significantly higher quality than that produced by its 3D counterpart, due to the vast difference in the amount of labeled training data they have been exposed to. To mitigate this, a novel triplet loss was developed that makes use of unlabeled test data in a transductive setting. The utility of this method was demonstrated via an extensive set of experiments that showed significant benefit in the 2D domain and established state-of-the-art results in the 3D domain for ZSL and GZSL tasks.

{\small
\bibliographystyle{ieee}
\bibliography{final_sup}
}

\clearpage
\pagebreak

\section*{Supplementary Material}


\maketitle
\ifwacvfinal\thispagestyle{empty}\fi

In this supplementary material, we further assess our proposed method with additional quantitative and qualitative evaluations. In the quantitative evaluation section, we evaluate \textbf{(1)} the effect of the batch size on 3D Zero-Shot Learning (ZSL) using ModelNet10, \textbf{(2)} the effect of using a different point cloud architecture, EdgeConv~\cite{Article24}, and \textbf{(3)} the effect of using the experimental protocol for Generalized Zero-Shot Learning (GZSL) proposed by Song \etal \cite{Song2018TransductiveUE}. In the qualitative evaluation section, we show success and failure cases on unseen classes from ModelNet10.

\section{Additional Quantitative Evaluation}
\subsection{Batch Size}

\begin{figure}[!h]\centering
\includegraphics[width=1\linewidth,trim=0cm 0cm 0cm 0cm, clip]{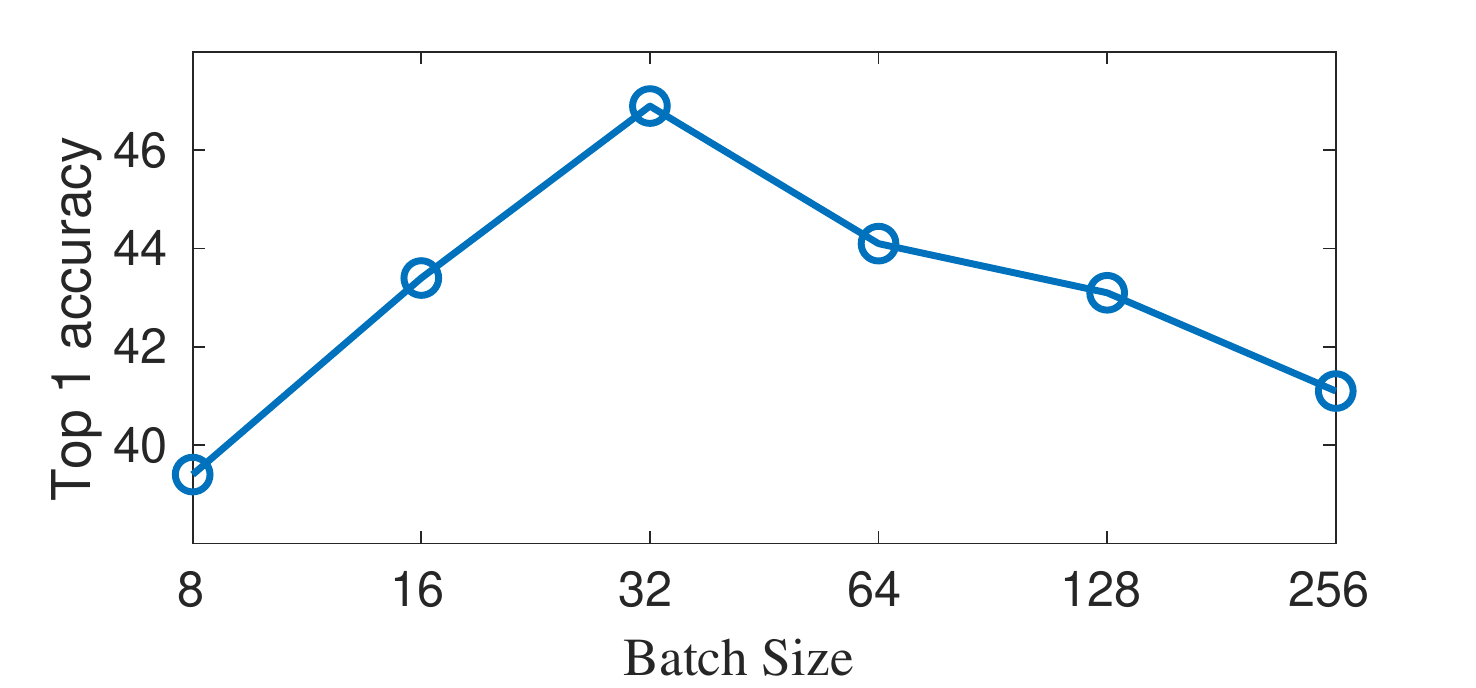}
\caption{Top-1 accuracy on the ModelNet10 dataset as the batch size varies.}
\label{fig:batch_size}
\end{figure}

In this experiment, we evaluate the effect of the batch size on the accuracy of our proposed method for the 3D ModelNet10 dataset. As can be seen in Figure~\ref{fig:batch_size}, the size of the batch has a significant impact on the performance, with the best performance on this dataset being achieved at a batch size of 32.

\subsection{Point Cloud Architecture}

\begin{figure*}
\centering
\includegraphics[width=1\linewidth,trim=0cm 0cm 0cm 0cm, clip]{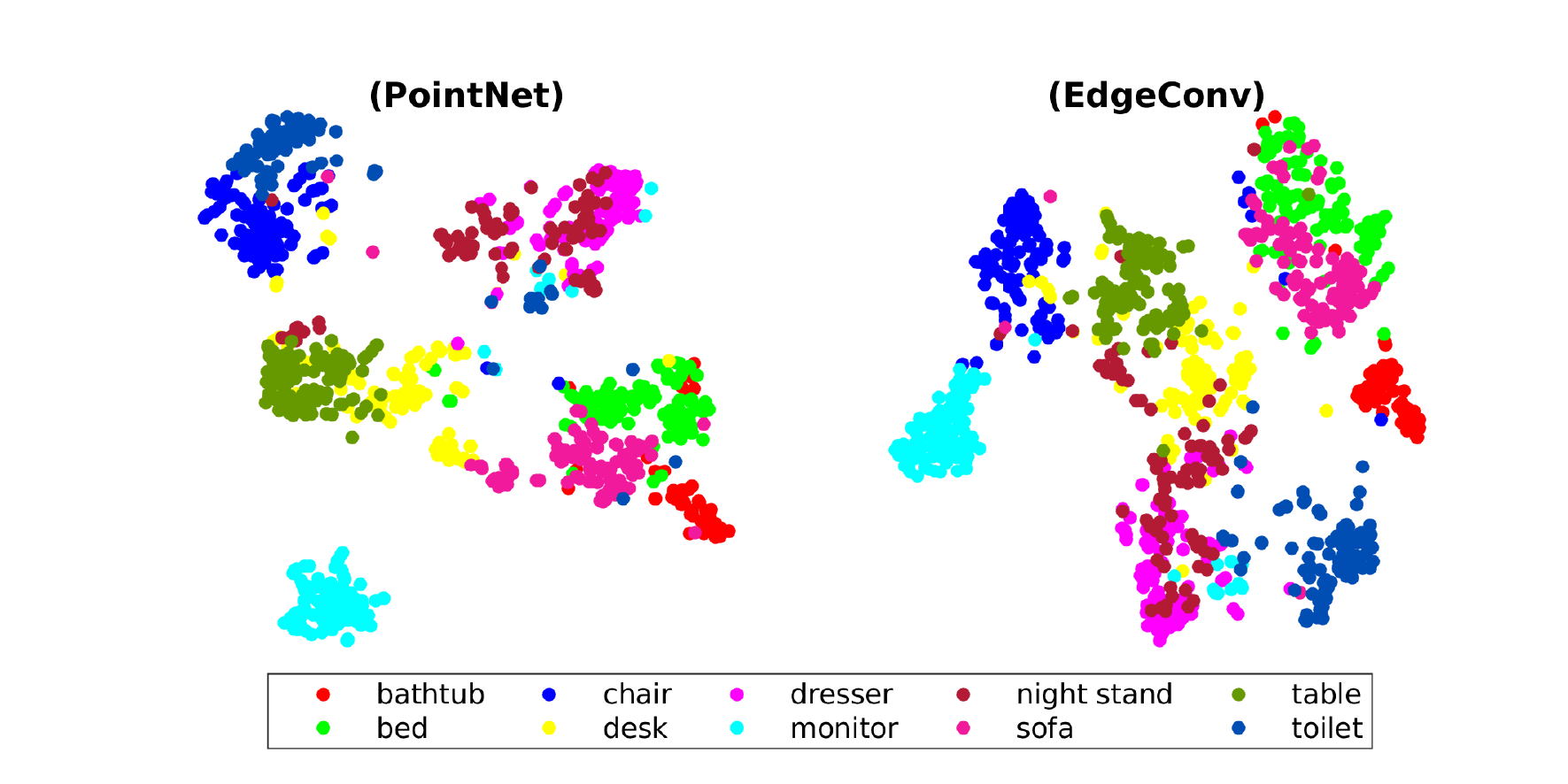}
\caption{2D tSNE~\cite{tSNE_van2014} visualization of unseen point cloud feature vectors (circles) based on (a) PointNet (b) EdgeConv on ModelNet10. The unseen point cloud features are clustered similarly in both PointNet and EdgeConv, despite EdgeConv performing better than PointNet on the task of supervised point cloud classification.}
\label{fig:point_cloud_architecture}
\end{figure*}

In this paper, we used PointNet \cite{Article1} as the backbone point cloud architecture in our 3D experiments. However, while PointNet is one of the first works that has been proposed for point cloud classification using deep learning, there are many other methods ~\cite{Article1,Article2,Article24,Article27,Article28,Article29,Xie_2018_CVPR,DBLP:journals/corr/abs-1712-06760,su18splatnet,8658405} which were introduced later and tend to achieve better performance for supervised 3D point cloud classification. Here, we compare PointNet with EdgeConv \cite{Article24} to study the effect of using a more advanced point cloud architecture for the task of 3D ZSL classification. In supervised 3D point cloud classification, EdgeConv achieves 92.2\% accuracy on ModelNet40 while PointNet achieves 89.2\%. In this additional experiment, we use ModelNet10 as the unseen set to compare those two methods. As shown in Table~\ref{table:ZSL_3D_compare}, both PointNet and EdgeConv achieve similar performance. We would expect to see some improvement when using EdgeConv since it works better in the case of supervised classification. In Figure \ref{fig:point_cloud_architecture}, it can be seen however that both PointNet and EdgeConv cluster unseen point cloud features similarly and imperfectly. This again shows the difficulty of the ZSL task on 3D data where there are a lack of good pretrained models. 

\begin{table}[!t]\centering
\newcolumntype{C}{>{\centering\arraybackslash}X}
\setlength{\tabcolsep}{4pt}
\begin{tabularx}{\columnwidth}{l C C C}\hline
  Method & ModelNet10 & McGill & SHREC2015 \\\hline
  PointNet \cite{Article1} & \textbf{46.9} & \textbf{21.7} & \textbf{13.0} \\ 
  EdgeConv \cite{Article24} & 45.2 & 20.6 & \textbf{13.0}\\
   \hline
\end{tabularx}
\vspace{0pt}
\caption{ZSL results on the 3D ModelNet10 \cite{Article10}, McGill \cite{Article49}, and SHREC2015 \cite{Article48} datasets using different point cloud architecture, PointNet and EdgeConv.}
\label{table:ZSL_3D_compare}
\end{table}

\subsection{QFSL's Generalized ZSL Evaluation Protocol}

In this experiment, we evaluate the effect of using a different evaluation protocol for the GZSL experiments, as proposed by Song \etal~\cite{Song2018TransductiveUE}. Under this protocol, the unlabeled data, which consists of seen and unseen instances, is divided into halves, and two models are trained. In each model, half of unlabeled data is used for training and the other half for testing. The final performance is calculated by averaging the performance of these two models. The authors suggest that this allows for fairer evaluation, although it is an imperfect solution.
Nonetheless, we show in Table~\ref{table:GZSL_3D_compare_QFSL_setting} for the ModelNet10 dataset that our method performs better than QFSL with respect to all accuracy measures under both this protocol and the original protocol from our paper. In fact, both methods perform better under this different protocol, which suggests that splitting the unlabeled data in this way makes the task easier. As a result, we use our more conservative GZSL evaluation protocol in the main paper.

\begin{table}[!t]\centering
\newcolumntype{C}{>{\centering\arraybackslash}X}
\setlength{\tabcolsep}{4pt}
\begin{tabularx}{\columnwidth}{l C C C}\hline
  Method &  $\Acc_{s}$ &  $\Acc_{u}$ & HM  \\\hline
  QFSL \cite{Song2018TransductiveUE} & 58.1 / 68.2 & 21.8 / 24.3 & 31.7 / 35.6 \\ 
  Ours  & \textbf{74.6} / 72.0 & 23.4 / \textbf{29.2} & 35.6 / \textbf{41.5} \\
   \hline
\end{tabularx}
\vspace{0pt}
\caption{GZSL results on the 3D ModelNet10 dataset \cite{Article10} under evaluation protocols (A) / (B), where (A) is the evaluation protocol from our paper and (B) is the protocol proposed by Song \etal \cite{Song2018TransductiveUE}. We report the top-1 accuracy (\%) on seen classes ($\Acc_s$) and unseen classes ($\Acc_u$) for each method, as well as the harmonic mean (HM) of both measures.}
\label{table:GZSL_3D_compare_QFSL_setting}
\end{table}

\section{Qualitative Evaluation}

\begin{figure*}
\centering
\includegraphics[width=0.8\linewidth,trim=0cm 0cm 0cm 0cm, clip]{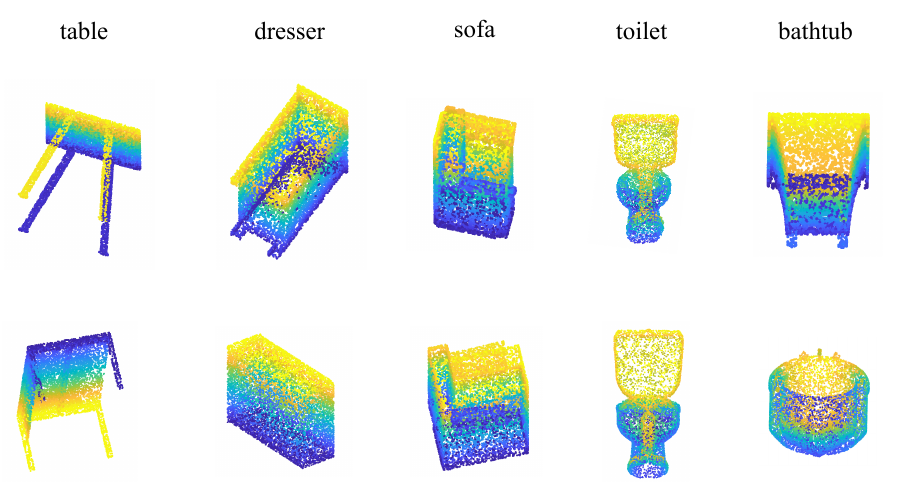}
\caption{Visualization of five classes from the ModelNet10 dataset with examples of correctly classified point clouds.}
\label{fig:success}
\end{figure*}

\begin{figure*}
\centering
\includegraphics[width=0.8\linewidth,trim=0cm 0cm 0cm 0cm, clip]{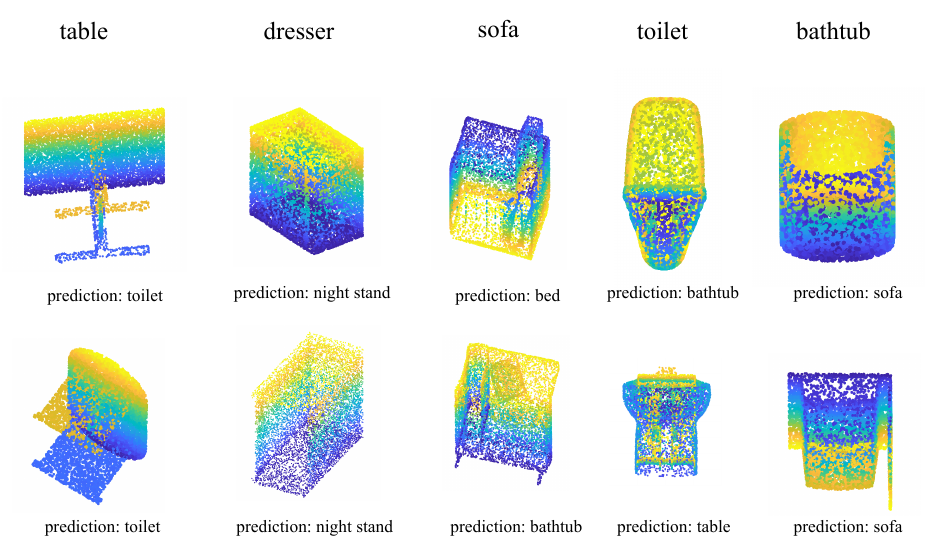}
\caption{Visualization of five classes from the ModelNet10 dataset with examples of incorrectly classified point clouds. The predicted classes are shown below each model.}
\label{fig:failure}
\end{figure*}

In this section, we visualize five unseen classes from the ModelNet10 dataset with examples where our method correctly classified the point cloud, shown in Figure \ref{fig:success}, and examples where it incorrectly classified the point cloud, shown in Figure \ref{fig:failure}. The network appears to be providing incorrect predictions for mostly hard examples, those that are quite different from standard examples in that class, or where the classes overlap in their geometry, such as dresser and night stand.

\end{document}